# Large Language Model Powered Automated Modeling and Optimization of Active Distribution Network Dispatch Problems

Xu Yang, *Graduate Student Member, IEEE*, Chenhui Lin, *Member, IEEE*, Yue Yang, *Member IEEE*,
Qi Wang, *Member, IEEE*, Haotian Liu, *Member, IEEE*, Haizhou Hua, and Wenchuan Wu, *Fellow, IEEE*

*Abstract*—The increasing penetration of distributed energy resources into active distribution networks (ADNs) has made effective ADN dispatch imperative. However, the numerous newly-integrated ADN operators, such as distribution system aggregators, virtual power plant managers, and end prosumers, often lack specialized expertise in power system operation, modeling, optimization, and programming. This knowledge gap renders reliance on human experts both costly and time-intensive. To address this challenge and enable intelligent, flexible ADN dispatch, this paper proposes a large language model (LLM) powered automated modeling and optimization approach. First, the ADN dispatch problems are decomposed into sequential stages, and a multi-LLM coordination architecture is designed. This framework comprises an Information Extractor, a Problem Formulator, and a Code Programmer, tasked with information retrieval, optimization problem formulation, and code implementation, respectively. Afterwards, tailored refinement techniques are developed for each LLM agent, greatly improving the accuracy and reliability of generated content. The proposed approach features a user-centric interface that enables ADN operators to derive dispatch strategies via simple natural language queries, eliminating technical barriers and increasing efficiency. Comprehensive comparisons and end-to-end demonstrations on various test cases validate the effectiveness of the proposed architecture and methods.

## I. INTRODUCTION

As an increasing number of distributed energy resources (DERs) are integrated into the distribution networks, the distribution networks are progressively evolving into active distribution networks (ADNs) [1], [2]. The coupling of active and reactive power, along with complex bidirectional power flows, has made traditional passive control strategies much less reliable [3]. As a result, the ADN dispatch has been proposed, which coordinates these DERs as well as other controllable devices within the ADN to enhance the overall safety and economic efficiency of the distribution system [4], [5].

Apart from that, with the rapid development of smart grid technologies and electricity markets, a variety of emerging operators, including distribution system aggregators, virtual power plant managers, and end prosumers, are joining the ADNs, becoming the main forces in ADN dispatch [6], [7]. These newly-integrated operators may have limited expertise in power system operation, modeling, optimization, and programming. At the same time, the expense and effort of hiring human experts specifically for dispatch purposes are too high for these operators to afford. Therefore, although the access barriers have been greatly diminished, efficient dispatch remains a significant challenge for these newly-integrated operators. These ADN operators call for a more flexible and intelligent automated dispatch approach, which not only meets their diverse requirements but also autonomously generates dispatch strategies, thus reducing the complexity of their entire dispatch process.

In recent years, advancements in large language models (LLMs) have provided a promising solution for automated dispatch in ADNs [8]. On the one hand, based on the vast amount of pre-training data, existing LLMs have already embedded certain domain knowledge concerning power systems, which can partially compensate for the deficiencies of the ADN operators [9]. On the other hand, based on the transformer architecture and billions of parameters, LLMs are able to capture the relationships between sentences, thereby facilitating comprehension, reasoning, and in-context learning to address diverse dispatch scenarios [10]. Finally, the inherent natural language processing capability of LLMs enables them to deliver a friendly interface for the operators, which further lowers the barrier to ADN dispatch.

Consequently, the accelerated evolution of LLMs has also brought about related research in the field of power systems [11]-[16]. For example, researchers in [11], [12] have introduced LLMs to perform power system simulations, which can generate simulation code based on external knowledge bases, thus little human intervention is required. Researchers in [13], [14] have explored possible LLM applications in the power system, such as risk recognition, document analysis, correlation analysis, and load forecasting. Also, researchers in [15] have proposed a benchmark to assess LLM performance in the power sector. And some senior

This work was supported in part by the Beijing Natural Science Foundation under Grant L243003 and the National Science Foundation of China under Grant U24B6009 *(Corresponding author: Wenchuan Wu).*

X. Yang, C. Lin, H. Liu, H. Hua, and W. Wu are with the State Key Laboratory of Power Systems, Department of Electrical Engineering, Tsinghua University, Beijing 100084, China.

Y. Yang is with the State Key Laboratory of High-Efficiency and High-Quality Conversion for Electric Power, Hefei University of Technology, Hefei 230009, China.

Q. Wang is with the Hong Kong Polytechnic University, Hong Kong, China.

scholars in [16] have discussed potential threats of applying LLMs to the power system and provided some research directions.

However, most existing LLM applications have not tackled dispatch problems—a critical challenge for realizing efficient ADNs. While studies in [17], [18] integrate LLMs with reinforcement learning (RL) for optimal power flow and energy management in ADNs, these approaches rely on separate RL agents to execute modeling and optimization. Moreover, the design of the RL agents also requires substantial domain knowledge and restricts their functionality to single, predefined tasks, limiting scalability across diverse ADN dispatch scenarios and problem variants.

The authors contend that the primary hurdle in applying LLMs to ADN dispatch stems from the intrinsic characteristics of dispatch itself. Unlike the previously discussed problems solvable via discrete LLM capabilities, ADN dispatch exhibits the following distinguishing traits:

*1) High complexity.* While ADN dispatch problems are linguistically straightforward to describe (e.g., *"Minimize tomorrow's operational costs."*), their underlying modeling and optimization complexities—characterized by multi-objective functions, high-dimensional decision spaces, and intricate constraints (especially with DER integration)—pose significant challenges for LLMs. This contrasts sharply with simpler tasks like risk recognition, where uploading a photo to a multimodal LLM with a query like *"What risks are present?"* yields direct, actionable outputs. However, for dispatch problems, the gap between natural language queries and optimal solutions widens due to its complexity.

*2) Versatility and diversity.* Due to the diverse operating scenarios, adaptable devices, and complex operational constraints in ADNs, the objectives, conditions, requirements, and expressions of dispatch requests are highly diverse and flexible, which may also involve some colloquial or vague words. Despite this ambiguity in inputs, dispatch problem solutions must adhere to strict standards of accuracy, safety, and reliability—requirements far more critical than in tasks like document analysis, where structured inputs and non-catastrophic error margins prevail. Unlike non-critical LLM applications, ADN dispatch demands a reliable bridge between ambiguous natural language and precise computational models to ensure the grid's safety and operational integrity. How to extract effective information from dispatch requests and transform it into structured content for LLMs to analyze and process is a critical step in automated dispatch.

*3) Dependence on extensive domain knowledge.* Solving ADN dispatch problems requires specialized, multidisciplinary domain knowledge spanning grid operation, mathematical modeling, optimization theory, and programming expertise. Unlike tasks like load forecasting, where large time-series models can leverage inherent data patterns, ADN dispatch often involves new operational scenarios, novel case formats, or updated modeling languages that may not be represented in LLM pre-training corpora. How to incorporate necessary domain knowledge into the content generated by LLMs is a key challenge in automated dispatch.

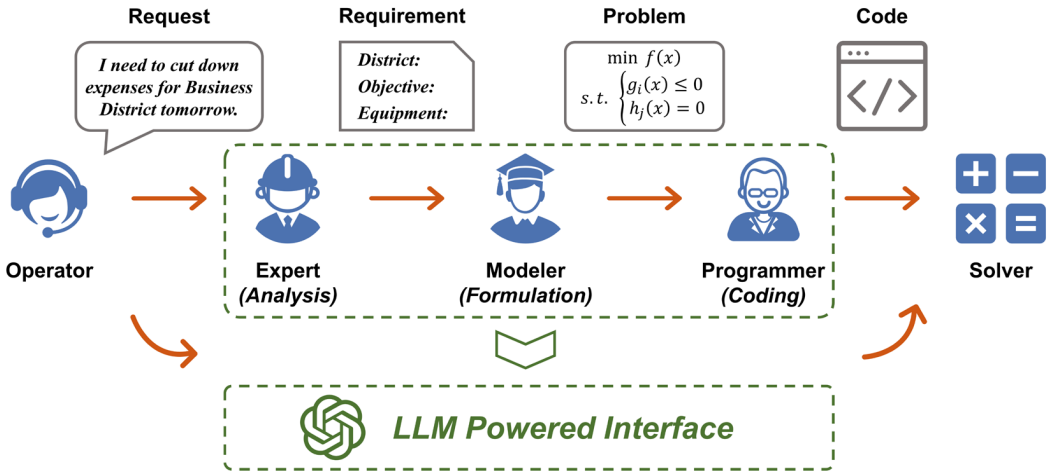

**Fig. 1.** Solution process of ADN dispatch problems by human experts.

Facing these difficulties, our LLM powered automated modeling and optimization method in this paper draws inspiration from how human experts tackle dispatch problems. As shown in Fig. 1, the solution of human experts typically consists of the following stages: First, the ADN operator provides a dispatch request, which is then passed to the experienced dispatch expert for analysis, converting the natural language-based request into structured dispatch requirements, including information concerning the district, objective, equipment, and so on. Subsequently, the modeler formulates these requirements as a constrained optimization problem, transforming the structured dispatch requirements into a unified math format. Afterwards, the programmer uses a modeling language to describe the constrained optimization problem again, translating the math format into an executable code format. Finally, the executable code is submitted to a commercial solver (Gurobi [19], CPLEX [20], COPT [21], etc.) to obtain the final dispatch strategies.

Following the problem-solving methodology of human experts, we decompose the resolution of dispatch problems into similar steps and propose a multi-LLM coordination architecture in this paper, including an LLM powered Information Extractor, an LLM powered Problem Formulator, and an LLM powered Code Programmer, which correspond to the above expert, modeler, and programmer, respectively. In this architecture, each LLM agent is responsible for only one subtask, simplifying their workload and reducing the required domain knowledge for each LLM agent, thereby enhancing the overall intelligence and reliability.

Furthermore, in order to manage the versatility and diversity of dispatch requests and requirements, as well as supplement the domain knowledge that existing LLMs may lack, we develop tailored enhancement methods for each LLM agent to assist in generating accurate outputs, including prompt methods, multi-round dialogues, few-shot learning [22], chain-of-thought (CoT) reasoning [23], and retrieval-augmented generation (RAG) [24]. Finally, to validate the proposed multi-LLM coordination architecture and enhancement methods, we conduct end-to-end evaluations and ablation studies, demonstrating the rationality and effectiveness. The main contributions of this paper can be summarized as follows:

1) **Multi-LLM coordination architecture.** We decompose the ADN dispatch problems into several steps and propose a multi-LLM coordination architecture. This architecture streamlines the transformation from natural language requests to optimized dispatch strategies. It comprises three specialized LLM agents, including Information Extractor, Problem Formulator, and Code Programmer. Information Extractor retrieves essential information from dispatch requests, such as district, dispatch objective, available equipment, and additional constraints. Using the retrieved information, Problem Formulator formalizes the dispatch request as a constrained optimization problem in math format. Based on the math format, Code Programmer translates it into executable code, which is then processed by the solver to obtain the final results.

2) **Prompt methods for Information Extractor.** To ensure information extraction accuracy, we design a series of prompt methods for ADN dispatch, comprising role/task/environment descriptions, output formats, few-shot examples, and CoT guidance. These methods apply uniformly to all LLM agents, i.e., Information Extractor, Problem Formulator, and Code Programmer. A core function of the Information Extractor is to translate diverse, colloquial dispatch requests into a structured schema, providing consistent inputs for downstream LLMs.

3) **Multi-round dialogue method for Problem Formulator.** Constructing a complete constrained optimization problem directly from extracted data remains challenging for LLMs due to complicated constraints and multi-domain complexity. We address this issue via a multi-round dialogue framework, where the Problem Formulator iteratively defines objective functions, layers equipment/power flow/additional constraints, and applies convex relaxation in sequential stages. It incrementally fills in the relevant domain knowledge, avoiding hallucinations or information loss in LLMs.

4) **External knowledge enhanced methods for Code Programmer.** While LLMs exhibit strong code generation capabilities, adapting to specialized modeling languages and case formats remains critical. To ensure accurate code generation, the Code Programmer is augmented with external knowledge modules, including annotated case format specifications and our domain-specific modeling language PyOptInterface [25]. Additionally, a novel RAG-assisted few-shot learning method is proposed, which dynamically retrieves and embeds semantically similar examples (based on mathematical problem structure) into prompts, enhancing contextual relevance.

## II. PRELIMINARIES

### A. Active Distribution Network Dispatch Problems

As a typical tree-structured network, if we define the power flow direction as from the root node to leaf nodes, the power flow of an ADN can be described by the Distflow model [26]:

$$P_{j,t} = \sum_{k:j\to k} P_{jk,t} - \sum_{i:i\to j}(P_{ij,t} - r_{ij}l_{ij,t}), \forall j \in \mathcal{N} \quad (1)$$

$$Q_{j,t} = \sum_{k:j\to k} Q_{jk,t} - \sum_{i:i\to j}(Q_{ij,t} - x_{ij}l_{ij,t}), \forall j \in \mathcal{N} \quad (2)$$

$$v_{j,t} = v_{i,t} - 2(r_{ij}P_{ij,t} + x_{ij}Q_{ij,t}) + (r_{ij}^2 + x_{ij}^2)l_{ij,t}, \forall ij \in \mathcal{E} \quad (3)$$

$$l_{ij,t} = \frac{P_{ij,t}^2 + Q_{ij,t}^2}{v_{i,t}}, \forall ij \in \mathcal{E} \quad (4)$$

$$v_{0,t} = 1.0 \quad (5)$$

where $\mathcal{N}$ and $\mathcal{E}$ are the collections of nodes and branches in the ADN; $P_{j,t}$ and $Q_{j,t}$ are the active and reactive power injection at bus $j$, time $t$; $P_{ij,t}$ and $Q_{ij,t}$ are the active and reactive power flow in branch $ij$, time $t$; $r_{ij}$ and $x_{ij}$ are the resistance and reactance of branch $ij$; $v_{j,t}$ is the squared voltage magnitude at bus $j$, time $t$; $l_{ij,t}$ is the squared current magnitude in branch $ij$, time $t$; eq. (5) indicates the root node voltage magnitude. Since eq. (4) is a non-convex constraint describing the relationship of $P_{ij,t}$, $Q_{ij,t}$, $v_{i,t}$, and $l_{ij,t}$, it should be relaxed as a second-order cone constraint:

$$\left\| \begin{array}{c} 2P_{ij,t} \\ 2Q_{ij,t} \\ l_{ij,t} - v_{i,t} \end{array} \right\|_2 \leq l_{ij,t} + v_{i,t}, \forall ij \in \mathcal{E} \quad (6)$$

Common equipment in an ADN includes diesel generator (DG), battery energy storage system (BESS), photovoltaic (PV), and static var compensator (SVC). If all of the above mentioned equipment is available, the active and reactive power injection at bus $i$, time $t$ can be expressed as follows:

$$P_{i,t} = P_{i,t}^{DG} + P_{i,t}^{BESS,dis} - P_{i,t}^{BESS,cha} + P_{i,t}^{PV} - P_{i,t}^{load}, \forall i \in \mathcal{N} \quad (7)$$

$$Q_{i,t} = Q_{i,t}^{DG} + Q_{i,t}^{PV} + Q_{i,t}^{SVC} - Q_{i,t}^{load}, \forall i \in \mathcal{N} \quad (8)$$

where $P_{i,t}^{DG}$ and $Q_{i,t}^{DG}$ are the active and reactive power generated by DG at bus $i$, time $t$; $P_{i,t}^{BESS,dis}$ and $P_{i,t}^{BESS,cha}$ are the discharging and charging power of BESS at bus $i$, time $t$; $P_{i,t}^{PV}$ and $Q_{i,t}^{PV}$ are the active and reactive power generated by PV at bus $i$, time $t$; $Q_{i,t}^{SVC}$ is the reactive power compensated by SVC at bus $i$, time $t$; $P_{i,t}^{load}$ and $Q_{i,t}^{load}$ are the active and reactive load at bus $i$, time $t$.

As for constraints for the above equipment, DG constraints are presented in eq. (9)-eq. (11), where $P_{min}^{DG}$ and $P_{max}^{DG}$ are the lower limit and upper limit of DG active power generation; $Q_{min}^{DG}$ and $Q_{max}^{DG}$ are the lower limit and upper limit of DG reactive power generation; $R_{max}^{DG}$ is the ramp rate limit of DG.

$$P_{min}^{DG} \leq P_{i,t}^{DG} \leq P_{max}^{DG}, \forall i \in \mathcal{N} \quad (9)$$

$$Q_{min}^{DG} \leq Q_{i,t}^{DG} \leq Q_{max}^{DG}, \forall i \in \mathcal{N} \quad (10)$$

$$-R_{max}^{DG} \leq P_{i,t}^{DG} - P_{i,t-1}^{DG} \leq R_{max}^{DG}, \forall i \in \mathcal{N} \quad (11)$$

BESS constraints are presented in eq. (12)-eq. (18), where $P_{max}^{BESS,dis}$ and $P_{max}^{BESS,cha}$ are the discharging and charging power limit of BESS; $SOC_{i,t}^{BESS}$ is the state of charge (SOC) of BESS at bus $i$, time $t$; $SOC_{min}^{BESS}$ and $SOC_{max}^{BESS}$ are the lower limit and upper limit of BESS SOC; $\eta$ is the discharging/charging efficiency; $\Delta t$ is the interval between two timesteps; $SOC_{init}^{BESS}$ is the initial BESS SOC; $SOC_{i,0}^{BESS}$ and $SOC_{i,T-1}^{BESS}$ are the initial and final BESS SOC of the dispatch process; $T$ is the length of the dispatch process.

$$0 \leq P_{i,t}^{BESS,dis} \leq P_{max}^{BESS,dis}, \forall i \in \mathcal{N} \quad (12)$$

$$0 \leq P_{i,t}^{BESS,cha} \leq P_{max}^{BESS,cha}, \forall i \in \mathcal{N} \quad (13)$$

$$P_{i,t}^{BESS,dis} \times P_{i,t}^{BESS,cha} = 0, \forall i \in \mathcal{N} \quad (14)$$

$$SOC_{min}^{BESS} \leq SOC_{i,t}^{BESS} \leq SOC_{max}^{BESS}, \forall i \in \mathcal{N} \quad (15)$$

$$SOC_{i,t}^{BESS} = SOC_{i,t-1}^{BESS} - \frac{P_{max}^{BESS,dis} \Delta t}{\eta} + \eta P_{max}^{BESS,cha} \Delta t, \forall i \in \mathcal{N} \quad (16)$$

$$SOC_{i,0}^{BESS} = SOC_{init}^{BESS}, \forall i \in \mathcal{N} \quad (17)$$

$$SOC_{i,T-1}^{BESS} = SOC_{i,0}^{BESS}, \forall i \in \mathcal{N} \quad (18)$$

Eq. (14) enforces that a specific BESS cannot discharge and charge simultaneously. To relax this non-convex constraint, a binary variable $u_{i,t}^{BESS} \in \{0,1\}$ indicating the BESS status (1 for discharging, and 0 for charging) can be introduced. Then eq. (12)-eq. (14) are reformulated as:

$$0 \leq P_{i,t}^{BESS,dis} \leq P_{max}^{BESS,dis} \times u_{i,t}^{BESS}, \forall i \in \mathcal{N} \quad (19)$$

$$0 \leq P_{i,t}^{BESS,cha} \leq P_{max}^{BESS,cha} \times \left(1 - u_{i,t}^{BESS}\right), \forall i \in \mathcal{N} \quad (20)$$

PV constraints are presented in eq. (21), where $S_{max}^{PV}$ is the installed capacity.

$$\left(P_{i,t}^{PV}\right)^2 + \left(Q_{i,t}^{PV}\right)^2 \leq (S_{max}^{PV})^2, \forall i \in \mathcal{N} \quad (21)$$

SVC constraints are presented in eq. (22), where $Q_{max}^{SVC}$ is the upper limit of SVC reactive power compensation.

$$0 \leq Q_{i,t}^{SVC} \leq Q_{max}^{SVC}, \forall i \in \mathcal{N} \quad (22)$$

Common objectives of ADN dispatch problems can be roughly divided into two categories: day-ahead objectives and single-timestep objectives. Day-ahead objectives include minimizing operational cost eq. (23), minimizing power loss eq. (24), and minimizing voltage deviation eq. (25), where $\rho^{DG}$, $\rho^{BESS,dis}$, and $\rho^{BESS,cha}$ are corresponding cost coefficients; $\rho_0$ is the price of buying electricity. Single-timestep objectives include eliminating voltage violation eq. (26), and eliminating branch power violation eq. (27), where $\hat{t}$ is the current timestep.

$$\min \sum_{t=0}^{T-1} \left( \sum_{i \in \mathcal{N}} \rho^{DG} P_{i,t}^{DG} + \sum_{i \in \mathcal{N}} \rho^{BESS,dis} P_{i,t}^{BESS,dis} + \sum_{i \in \mathcal{N}} \rho^{BESS,cha} P_{i,t}^{BESS,cha} + \rho_0 P_{0,t} \right) \quad (23)$$

$$\min \sum_{t=0}^{T-1} \sum_{i \in \mathcal{N}} P_{i,t} \quad (24)$$

$$\min \sum_{t=0}^{T-1} \sum_{i \in \mathcal{N}} \left( v_{i,t} - 1.0 \right)^2 \quad (25)$$

$$\min \max_{i \in \mathcal{N}} \left( v_{i,\hat{t}} - 1.0 \right)^2 \quad (26)$$

$$\min \max_{ij \in \mathcal{E}} \left( P_{ij,\hat{t}}^2 + Q_{ij,\hat{t}}^2 \right) \quad (27)$$

To cope with non-convex objectives eq. (26) and eq. (27), we can introduce a new variable $z$ and new constraints eq. (28) and eq. (29) so that the objective can be reformulated as $\min z$.

$$z \geq \left( v_{i,\hat{t}} - 1.0 \right)^2, \forall i \in \mathcal{N} \quad (28)$$

$$z \geq \left( P_{ij,\hat{t}}^2 + Q_{ij,\hat{t}}^2 \right), \forall ij \in \mathcal{E} \quad (29)$$

For single-timestep objectives, these temporal coupling constraints, such as eq. (11) and eq. (15)-eq. (18) can be ignored in the optimization problem. In addition, if a certain type of equipment is unavailable in this district, such as not installed, under maintenance, or in failure, corresponding terms can be directly discarded from eq. (7), eq. (8), and eq. (23).

In practice, ADN operators also impose additional requirements based on local conditions, the most common of which are voltage safety constraints eq. (30) and branch power safety constraints eq. (31), where $V_{min}$ and $V_{max}$ are lower limit and upper limit of voltage magnitudes; $S_{max}^{brch}$ is the branch power capacity.

$$V_{min}^2 \leq v_{i,t} \leq V_{max}^2, \forall i \in \mathcal{N} \quad (30)$$

$$P_{ij,t}^2 + Q_{ij,t}^2 \leq (S_{max}^{brch})^2, \forall ij \in \mathcal{E} \quad (31)$$

All of the above modeling knowledge is incrementally fed into Problem Formulator, whose details are described in Section III. It should be noted that the above equations are common equipment, objectives, and operational requirements in ADNs. If there are further alternative modeling functions, they can be seamlessly integrated into prompts and dialogues following our proposed method.

### B. Large Language Model

In recent years, LLMs have made remarkable progress in the field of natural language processing. These LLMs are typically based on the transformer architecture and self-attention mechanisms to capture dependencies within text sequences, endowing them with strong comprehension, reasoning, and in-context learning capabilities [27]. Trained on enormous textual corpora and featuring large-scale parameters ranging from millions to billions, LLMs are able to internalize vast amounts of domain expertise. Moreover, LLMs are highly flexible and adaptive, capable of supporting multi-task execution and few-shot learning scenarios, making them one of the core technologies of modern artificial intelligence.

On the application front, LLMs have already been widely adopted in various fields such as natural language comprehension, information retrieval, and translation, driving the expansion of artificial intelligence into a broader range of real-world tasks. Beyond these traditional applications, leveraging their strong reasoning capabilities and broad domain knowledge, some researchers are beginning to employ LLMs as policy models to perform decision-making tasks and action planning, such as robotic control [28], [29], computer games [30], and code programming [31].

However, in terms of power system applications, most of the existing studies are centered on applying LLMs to tasks like forecasting and document analysis. Despite being a typical decision-making task in the power system, LLM for dispatch problems has not been thoroughly studied or widely applied. Our proposed "natural language to executable code" LLM interface and automated modeling and optimization method are a viable and promising pathway for integrating LLMs into dispatch problems, contributing improved decision-making intelligence and operational efficiency.

## III. Methods

### A. Multi-LLM Coordination Architecture

In this study, we decompose ADN dispatch problem-solving process into multiple sequential steps and propose a multi-LLM coordination architecture accordingly. The proposed architecture comprises three specialized LLM agents, denoted as Information Extractor, Problem Formulator, and Code Programmer. Detailed descriptions of each agent are as follows:

1) *Information Extractor:* Since dispatch requests are all based on natural language and exhibit a high degree of flexibility and diversity, the Information Extractor distills and organizes the essential information from these requests to ensure consistency and accuracy for downstream LLM agents. Input of the Information Extractor is ADN dispatch requests expressed in natural language and its output is structured requirements, including the controlled district, dispatch objective, available equipment, and additional constraints. Meanwhile, in order to maintain uniformity in the output format, all of the extracted information is wrapped in corresponding decorators. For example, the dispatch objective is enclosed with "<objective>" and "</objective>" tags for efficient parsing in later stages.

2) *Problem Formulator:* After extracting the information, it is still challenging for LLMs to directly map this information into executable code. Therefore, transforming the information into a constrained optimization problem in math format makes code generation process much easier. Based on this idea, the Problem Formulator constructs the corresponding optimization problem using the refined information from Information Extractor and our provided modeling knowledge. Input of the Problem Formulator is the structured requirements and its output is the constructed optimization problem in math format. Also, all of the non-convex objectives and constraints are relaxed using the methods described in Section II.

3) *Code Programmer:* To derive the final ADN dispatch strategies, coding is an indispensable step, since existing LLMs are unable to directly solve complex optimization problems. Input of the Code Programmer is the structured requirements and the constructed optimization problem and its output is the final code. After transforming the problem into executable code, it is passed to a commercial solver for resolution.

On the one hand, the proposed architecture provides a feasible and implementable "natural language to executable code" approach, offering a practical path for automated modeling and optimization of ADN dispatch problems. On the other hand, it reduces the workload and required knowledge for each LLM agent, improving the overall accuracy and reliability.

### B. Prompt Based Information Extractor

To ensure that the Information Extractor successfully distills useful information and maintains accurate and standardized output content, we develop corresponding system prompts for the LLM agent. As illustrated in Fig. 2, the prompts consist of the following components:

1) *Role description:* This component mainly describes the role of the LLM agent, that is, an expert in power system

operation and optimization, which encourages it to apply corresponding embedded knowledge in the conversations.

2) *Task description:* This component mainly describes the task of the LLM agent, which is to identify and extract valid information from ADN dispatch requests and organize it into the required format.

3) *Environment description:* This component contains external knowledge that describes the ADN environment faced by the LLM agent. It includes the introduction to districts, common dispatch objectives, available equipment, and typical additional constraints. This component should be described in detail and with precision, enabling the LLM agent to effectively understand and adapt to current scenario.

4) *Output format:* This component mainly describes the required output format. Additionally, since modeling different parts of the optimization problem relies on different information, we wrap the district, objective, equipment, and constraints in different decorators.

5) *Few-shot examples:* This component provides 3-5 examples to assist the LLM agent in performing imitation and few-shot learning. For the Information Extractor, a specific example includes a dispatch request and its corresponding structured output.

6) *CoT guidance:* CoT refers to breaking down a complex problem into smaller steps, allowing the LLM agent to think and generate outputs according to the required reasoning process, which enhances the precision of its answers.

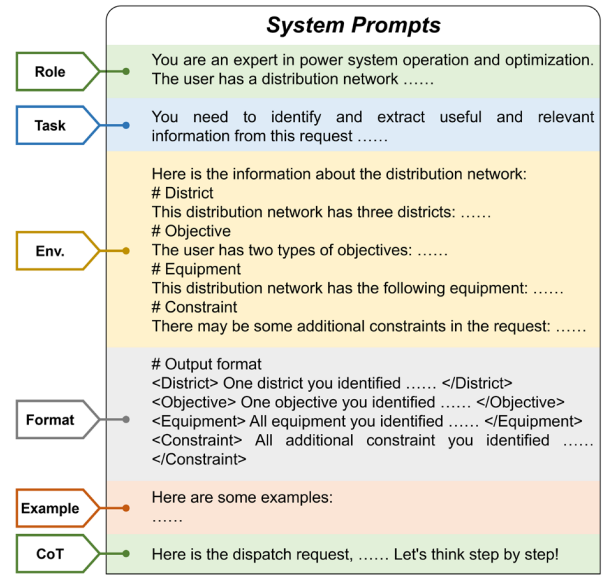

**Fig. 2.** Prompt methods for Information Extractor.

In summary, these system prompts are designed to use concise and accurate instructions to clearly address the key questions for the LLM agent, "Who are you?", "What should you do?", and "How should you do it?". Also, it should be emphasized that the prompts for the subsequent LLM agents will also contain elements like role description, task description, environment description, output format, few-shot examples, and CoT guidance. Therefore, to avoid repetition, the content of the prompts will not be introduced in the following subsections. Instead, the focus will be placed on the innovative enhancement methods. Complete prompts for all LLM agents can be found in the online supplementary file [32].

## C. Multi-Round Dialogue Based Problem Formulator

As shown in Fig. 3, after extracting information from the dispatch requests, the constrained optimization problem is built incrementally through multi-round dialogues, with each round focusing on a different part of the problem. The entire procedure proceeds as follows:

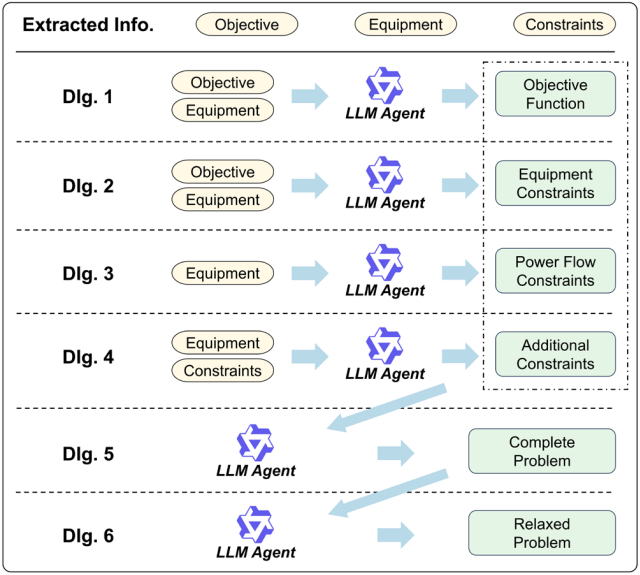

**Fig. 3.** Multi-round dialogues for Problem Formulator.

1) Dialogue 1: Based on the identified dispatch objective, the corresponding knowledge about the objective function is extracted from the modeling knowledge base and embedded into the prompts, enabling the LLM agent to write the math expression of the objective function. The equipment information is necessary in this step since the objectives like eq. (23) is relevant to available equipment.

2) Dialogue 2: Based on the identified available equipment, the corresponding knowledge about the equipment constraints is extracted and embedded into the prompts, enabling the LLM agent to write the math expression of the equipment constraints. Also, the objective is necessary in this step since day-ahead objectives will involve temporal coupling constraints while single-timestep objectives do not.

3) Dialogue 3: Based on the power flow knowledge, the LLM agent writes power flow functions like eq. (1)-eq. (5). Then, based on the available equipment, the LLM agent writes the power injection expressions like eq. (7) and eq. (8).

4) Dialogue 4: Based on the equipment and additional constraints, the LLM agent writes the additional constraints in math format.

5) Dialogue 5: After all of the above objective and constraints are constructed, we instruct the LLM agent to use a consistent set of symbols to organize the objective and constraints into a complete optimization problem.

6) Dialogue 6: Finally, based on the complete problem organized by the LLM agent, we guide it to apply convex relaxation techniques to relax possible non-convex objective and constraints, forming a solvable optimization problem.

This multi-round dialogue pattern allows modeling knowledge to be gradually incorporated into the LLM agent, reducing the complexity of the modeling process. It is also important to clarify that the conversation content with the LLM agent is predefined, only the corresponding parts need to be embedded, without the need for human intervention.

## D. External Knowledge Enhanced Code Programmer

The final step involves the LLM powered Code Programmer generating the executable code so that we can utilize the commercial solver to derive the final solutions. During the code generation process, it is highly likely that the LLM agent will encounter scenarios and modeling languages it has never seen before. This situation is quite common in real-world ADN applications, as there is no guarantee that the case formats and modeling languages of the ADN entity have been included in the pre-training corpus of the LLM. Therefore, it is necessary to provide sufficient external knowledge to support the code generation of the LLM agent. Apart from the complete optimization problem constructed by the Problem Formulator in the previous step, we also provide the following external knowledge:

1) Case format explanations: In this paper, the parameters (such as the number of BESS, generation cost coefficients), models (such as equipment locations), and data (such as PV and load profiles) of the case are stored in the form of a Python dictionary. In this part, we provide each key in the dictionary along with its corresponding explanations to facilitate the LLM's import of the test case.

2) PyOptInterface modeling language explanations: PyOptInterface is a novel Python-based modeling language. Compared to traditional modeling languages, its materials may not have been learned by existing LLMs, making it suitable for tests. In this part, we provide detailed explanations of common functions in PyOptInterface, including importing libraries, defining decision variables, creating constraints, setting up the objective function, and invoking the solver. Each function is accompanied by a thorough description of its usage to assist the LLM agent in generating code.

3) Few-shot examples: To reduce the difficulty of code generation, we also provide three examples for few-shot learning, each containing the math format of the optimization problem and the corresponding complete code.

In addition, considering the high diversity of dispatch problems, fixed examples are no longer suitable here. Instead, we propose a novel RAG-assisted few-shot learning method. Specifically, we have human experts pre-write the code and corresponding math formats for several dispatch optimization problems. Then using a text-embedding model to vectorize these math expressions and store them in a database. After receiving the math problem constructed by the Problem Formulator, we use the same text-embedding model to vectorize current problem and compute its cosine similarity with the problems in the database. For two n-dimensional vectors $A = (a_1, a_2, ..., a_n)$ and $B = (b_1, b_2, ..., b_n)$, their cosine similarity is defined as:

$$Cosine\ Similarity(A, B) = \frac{\sum_{i=1}^{n} a_i b_i}{\sqrt{\sum_{i=1}^{n} a_i^2} \times \sqrt{\sum_{i=1}^{n} b_i^2}} \quad (32)$$

Finally, three examples with the highest similarity are selected for few-shot learning. This ensures that regardless of how varied the problem is, appropriate examples remain available to reference.

## IV. NUMERICAL STUDIES

In this section, to validate the effectiveness of the proposed architecture and methods, we set up an ADN dispatch scenario and conduct comprehensive tests and numerical simulations. In this scenario, the ADN operator manages three districts: the valley district, the railway district, and the business district, which are represented by IEEE 33-bus [33], 69-bus [34], and 141-bus [35] distribution systems, respectively. Corresponding topologies, profiles, and parameters are provided in the supplementary file.

### A. Method and Case Settings

First, to verify the effectiveness of the proposed multi-LLM coordination architecture, we design ablation studies on each LLM agent within the architecture. "Full" represents the proposed approach, where all LLM agents in the architecture and enhancement methods are complete and included. "No-IE (No Information Extractor)" represents the absence of the Information Extractor, where dispatch requests are fed directly into the subsequent LLM agents without any processing. "No-PF (No Problem Formulator)" represents the absence of the Problem Formulator, where the final code is generated based on the extracted information instead of the optimization problem in math format. "No-IEPF (No Information Extractor and Problem Formulator)" represents the absence of both Information Extractor and Problem Formulator, where the Code Programmer solely writes the code based on the natural language dispatch requests.

Meanwhile, to demonstrate the proposed enhancement methods, we also design ablation studies targeting a specific method within the prompts. "No-EK (No External Knowledge)" represents that no external knowledge is provided, and all content generation is based only on the internal knowledge of the LLM. "No-FS (No Few-Shot Learning)" represents that no relevant examples are provided during the content generation process. "No-RAG" represents that although examples are provided, they are fixed and not retrieved dynamically using RAG technique based on the current dispatch request. The above methods are tested using a commercial LLM qwen-plus and an open-source LLM qwen2.5-72b. In addition, to assess the scalability of the proposed method and investigate how LLM parameter size affects its performance, we test the "Full" method on qwen2.5 LLMs with different parameter scales. Embedding model used in this paper is text-embedding-v3 provided by Aliyun. Configurations of the LLMs are provided in Table I.

For the case settings, we design 10 dispatch requests for each district for testing (30 requests in total), covering common objectives, equipment, and operational requirements described in Section II. Also, the designed requests simulate various operator tones, including colloquial, formal, brief, and comprehensive styles. All of the test requests are also provided in the supplementary file. Due to the stochasticity in the response of LLMs, we conducted three tests for each request using different random seeds. As a result, each method is evaluated 90 times, and the following results are the average values across these 90 tests.

TABLE I
CONFIGURATIONS OF LLMs

| Parameter | Value |
|---|---|
| Temperature | 0.6 |
| Top-p | 0.7 |
| Embedding Dimension | 1024 |
| Version (qwen-plus) | "2025-04-28" |

To quantitatively evaluate the performance of each method, we implement a scoring mechanism by human experts for the problem formulation results, dividing the full score of 100 into five 20-point parts: 1) objective function score; 2) equipment constraints score; 3) power flow constraints score; 4) additional constraints score; and 5) convexification score. For each of these five components, we follow the scoring criterion proposed in [12]:

1) $score = 20$ if all of the formulation requirements are fully satisfied;
2) $score = 10$ if part of the formulation requirements are satisfied, but include some errors or irrelevant information;
3) $score = 0$ if any requirement is not satisfied.

Although this scoring criterion is simple and may introduce some human bias, it still serves as a convenient and intuitive evaluation approach.

When evaluating the final generated code, in addition to the scoring mechanism, another important metric is their pass rates. Here, we utilize the "pass@k" criteria in the field of code generation, whose definition is as follows:

1) pass@1 denotes the probability that the generated code passes the test on the first attempt, i.e., the proportion of executable code among the 90 tests;
2) pass@3 denotes the probability of success within three attempts. For each of the 30 requests, if at least one out of the three generated code is executable, the request is considered passed.

Also, it should be noted that in this paper, the pass rates only measure the code's executability rather than correctness, as the latter is already evaluated through the aforementioned scoring mechanism. Thus, the pass rates reflect the LLM's programming capability and the method's comprehension of PyOptInterface modeling language. The detailed scores and pass/fail results of each test are also provided in the supplementary file.

### B. End-to-End Performance

First, in order to better visualize the proposed method and its performance, we illustrate an end-to-end example of "Full" method in Fig. 4, including output of the Information Extractor, optimization problem constructed by the Problem Formulator, code written by the Code Programmer, and the final ADN dispatch results of the code. LLM used in this example is qwen-plus. The optimization problem and code shown in the figure are direct outputs from the LLM agent without any human modifications. We only remove parts of the code for clarity and readability.

As shown in Fig. 4, in this example, the ADN operator proposes a dispatch request to minimize power loss in the valley district. The available devices include PV and SVC, and voltage safety constraints must be satisfied. Based on the request, the Information Extractor retrieves essential information within it and wraps the information in corresponding decorators. Afterwards, the information is sent to the Problem Formulator, based on which this LLM agent successfully constructs the corresponding optimization problem. Finally, the Code Programmer transforms the math format into executable code. At the bottom of the figure, we plot the voltage profiles of each bus before and after optimization. As can be seen, the voltage magnitudes before optimization exceed both the upper limit (peak PV generation at noon) and the lower limit, and the power loss is relatively high. By regulating the reactive generation of PV and SVC, the voltage is constrained within the safe range, and the power loss is reduced by 12.6%.

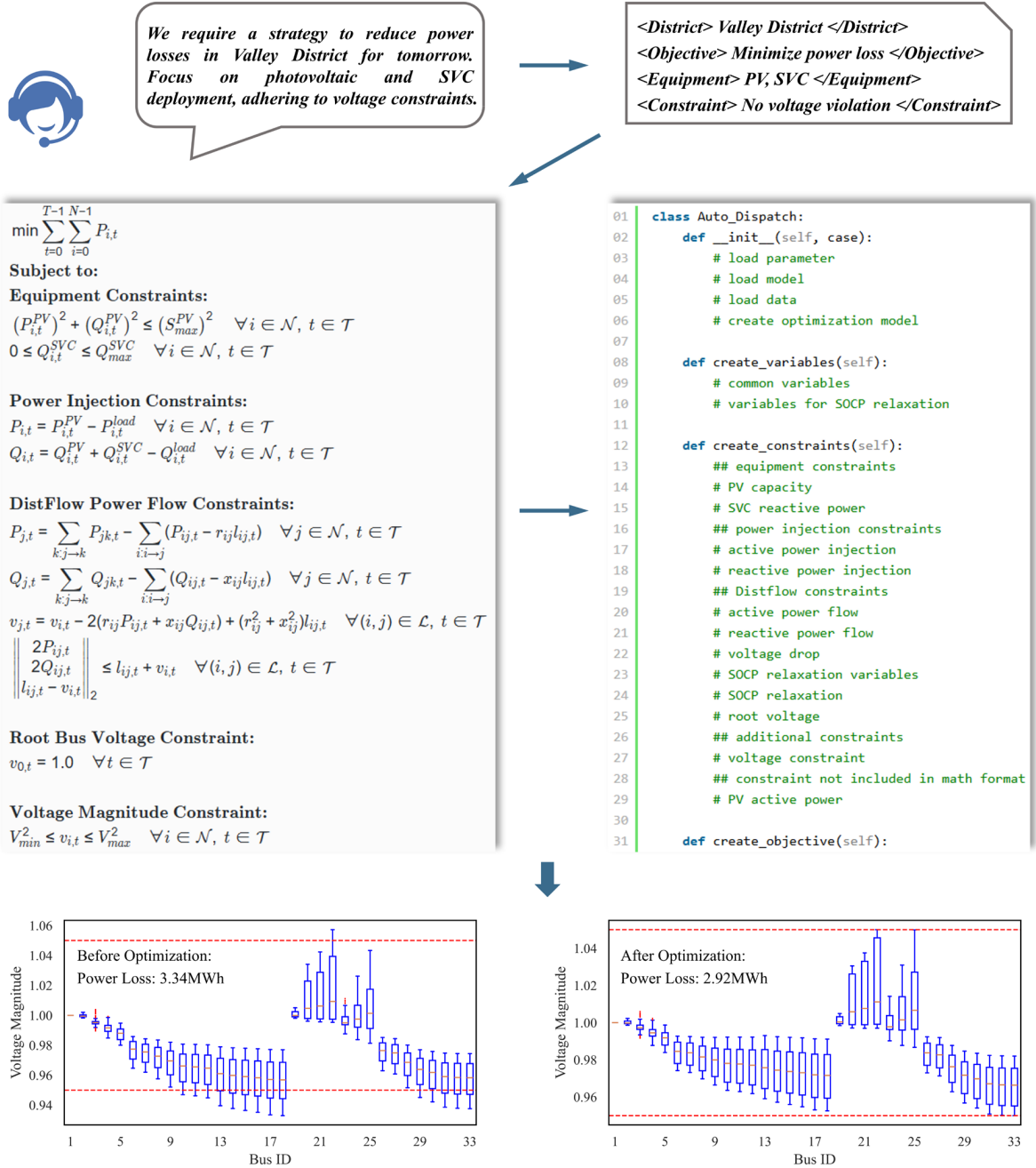

**Fig. 4.** Illustration of one end-to-end example of the proposed method.

## C. Ablation Studies

First, to visualize the performance of different methods on problem formulation and code programming, we plot their average scores in Fig. 5-Fig. 6, with blue indicating qwen-plus and red indicating qwen2.5-72b. As can be seen from the results, regardless of which LLM is used, "Full" achieves excellent performance close to the full score on both tasks, which validates the effectiveness of the proposed architecture and methods. Common errors of "Full" include symbol inaccuracies and adding/removing a term during modeling. Such errors are not related to the proposed method but rather stem from the LLM's inherent accuracy issues during content generation.

In the ablation studies concerning LLM agents, we first examine the performance of "No-IE" method. Although the Information Extractor only represents a relatively small function, its importance is apparent from the significant decrease in scores after the Information Extractor is removed. The reason lies in two aspects: 1) Information Extractor is provided with detailed ADN environment descriptions, enabling it to capture the most accurate and useful information compared to other LLM agents; 2) It offers structured outputs, which avoid exposing dispatch requests directly to subsequent agents. The absence of Information Extractor greatly increases the difficulty for other LLM agents. During the experiments, we find that the most frequent error of "No-IE" is misrecognition, specifically, the objectives and constraints do not match the requirements specified in the dispatch requests, which also demonstrates the necessity of Information Extractor in the proposed architecture. Compared to "Full" method, the removal of Problem Formulator in "No-PF" leads to a greater decline in performance. The underlying reason is that for LLMs, mapping formal math expressions into code is much simpler than directly translating natural language into code, which verifies the necessity of Problem Formulator. Based on the above analysis, it comes as no surprise that "No-IEPF" exhibits the poorest performance among the "Full", "No-IE", and "No-PF" methods.

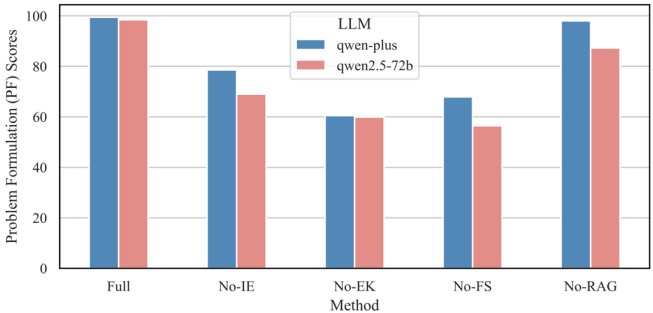

**Fig. 5.** Problem formulation (PF) scores of different methods.

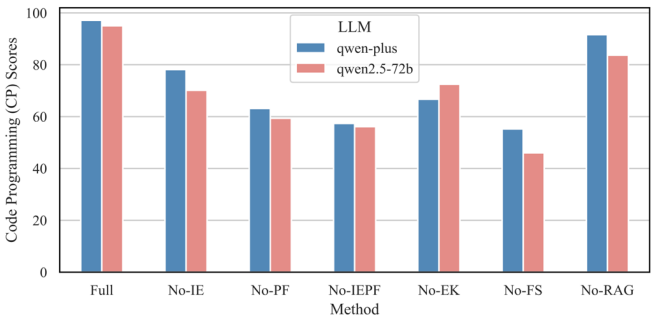

**Fig. 6.** Code programming (CP) scores of different methods.

As for the ablation studies concerning enhancement methods, it can be observed that the performance of "No-RAG" drops only slightly compared to the "Full" method. In particular, when the LLM is qwen-plus, its score is almost as high as "Full". This result suggests that despite the lack of the most similar examples, existing advanced LLMs like qwen-plus can still comprehend the task and generate logically sound results. In contrast, "No-EK" and "No-FS" achieve much lower scores. This highlights that on the one hand, relying solely on internal knowledge of LLM is inadequate, and external knowledge is essential to meet diverse requirements. On the other hand, a certain number of examples are crucial for few-shot learning, as they implicitly contain a wealth of domain knowledge. Based on these examples, LLMs can imitate desired outputs, which greatly simplifies the generation process and enhances the accuracy of results.

TABLE II
PASS RATES OF DIFFERENT METHODS

| Method | qwen-plus | | qwen2.5-72b | |
| --- | --- | --- | --- | --- |
| | pass@1 | pass@3 | pass@1 | pass@3 |
| Full | 0.98 | 1.00 | 0.93 | 1.00 |
| No-IE | 0.87 | 0.97 | 0.54 | 0.60 |
| No-PF | 0.98 | 1.00 | 0.88 | 1.00 |
| No-IEPF | 0.94 | 1.00 | 0.98 | 1.00 |
| No-EK | 0.37 | 0.70 | 0.48 | 0.77 |
| No-FS | 0.00 | 0.00 | 0.00 | 0.00 |
| No-RAG | 0.63 | 0.73 | 0.63 | 0.70 |

Second, we list the pass rates of different methods in Table II. As can be seen from the results, in the vast majority of tests, the "Full" method generates executable code on the first attempt, and its pass@3 reaches 100%, further confirming the validity of the proposed architecture and methods. Pass rates of "No-PF" and "No-IEPF" methods decline to some extent compared to "Full" method, and "No-IE" method delivers the worst performance among the first four methods. This is because "No-PF" and "No-IEPF" write the code directly based on dispatch requests, whereas the "No-IE" writes the code according to the math format. However, due to the lack of the Information Extractor, this math format may not be precise, which can lead to bugs or exceptions during code generation. Another interesting observation is the performance of "No-EK" and "No-FS". Although "No-EK" method does not provide external knowledge, it offers examples for LLM to imitate, resulting in pass@3 still higher than 70%. By contrast, despite offering detailed explanations on PyOptInterface, "No-FS" fails to produce executable code due to the lack of examples. This result provides compelling evidence that incorporating examples for few-shot learning is critically important for complex code generation tasks. As for the performance of "No-RAG", although this method achieves high scores in human expert evaluations, its code pass rates show a noticeable decline. This is due to the limited coverage of static examples, which fail to represent all possible scenarios, thereby increasing the probability of errors when the LLM agent encounters unrepresented cases. As a reminder, the pass rates only reflect code executability. During the experiments, we also observe that the main errors of "No-IE" and "No-RAG" are related to the usage of functions in PyOptInterface.

TABLE III
QWEN2.5 PERFORMANCE WITH DIFFERENT PARAMETER SIZES

| Parameter Size | PF Score | CP Score | pass@1 | pass@3 |
| --- | --- | --- | --- | --- |
| qwen2.5-72b | 98.4 | 95.0 | 0.93 | 1.00 |
| qwen2.5-32b | 95.0 | 92.0 | 0.98 | 1.00 |
| qwen2.5-14b | 91.4 | 88.2 | 0.91 | 1.00 |
| qwen2.5-7b | 66.2 | 65.8 | 0.76 | 0.97 |
| qwen2.5-3b | - | - | - | - |

Finally, we list the performance of qwen2.5 with different parameter sizes in Table III. It can be observed from the table that as the number of parameters decreases, the LLMs' performance gradually declines. Notably, when the parameter size reduces to 7b, there is a significant drop in performance. With the parameter size reduced to 3b, the LLM can no longer meet the requirements of problem formulation and code programming. Since dispatch is a relatively complex problem, a certain level of parameter size is required to support the performance of the LLM agent.

## V. Conclusion

To achieve automated modeling and optimization of ADN dispatch problems, we propose a multi-LLM coordination architecture that includes Information Extractor, Problem Formulator, and Code Programmer, along with corresponding enhancement methods for each LLM agent. This framework creates a natural language-to-code pipeline, which greatly alleviates the burden of ADN operators. Comprehensive test results validate the method's effectiveness. In addition, key experiments also reveal several critical LLM performance factors: complete external knowledge, appropriate examples, and sufficient model parameters.

This work advances LLM-driven ADN automation, lowering barriers for non-expert users to achieve efficient grid management. For ADN operators facing LLM deployment challenges, future research will explore lightweight model implementations to broaden applicability.